\documentclass[sigconf]{acmart}

\usepackage[utf8]{inputenc} % allow utf-8 input
\usepackage[T1]{fontenc}    % use 8-bit T1 fonts
\usepackage{hyperref}       % hyperlinks
\usepackage{url}            % simple URL typesetting
\usepackage{booktabs}       % professional-quality tables
\usepackage{amsfonts}       % blackboard math symbols
\usepackage{nicefrac}       % compact symbols for 1/2, etc.
\usepackage{microtype}      % microtypography
\usepackage{xcolor}         % colors
\usepackage{amsmath}

\usepackage{graphicx}
\usepackage{booktabs}
\usepackage{threeparttable}
\usepackage{array}
\usepackage{makecell}
\usepackage{caption}
\usepackage{subcaption}
\usepackage{multirow}
  
\newcommand \dataname {ZooplanktonBench}

\AtBeginDocument{%
  }

\copyrightyear{2025}
\acmYear{2025}
\setcopyright{cc}
\setcctype{by}
\acmConference[KDD '25]{Proceedings of the 31st ACM SIGKDD Conference on Knowledge Discovery and Data Mining V.2}{August 3--7, 2025}{Toronto, ON, Canada}
\acmBooktitle{Proceedings of the 31st ACM SIGKDD Conference on Knowledge Discovery and Data Mining V.2 (KDD '25), August 3--7, 2025, Toronto, ON, Canada}
\acmDOI{10.1145/3711896.3737395}
\acmISBN{979-8-4007-1454-2/2025/08}
\acmSubmissionID{dtbfb0064}

\begin{document}

%%
%% The "title" command has an optional parameter,
%% allowing the author to define a "short title" to be used in page headers.

% \title{
% ZooplanktonCV: A Benchmark Dataset for Zooplankton Recognition in Images and Videos}

\title{
ZooplanktonBench: A Geo-Aware Zooplankton Recognition and Classification Dataset from Marine Observations
}

%%
%% By default, the full list of authors will be used in the page
%% headers. Often, this list is too long, and will overlap
%% other information printed in the page headers. This command allows
%% the author to define a more concise list
%% of authors' names for this purpose.
% \author{%
%   Fukun Liu$^1$, \quad  Adam T Greer$^2$, \quad Gengchen Mai$^{3,4}$, \quad Jin Sun$^1$ \\
%   % \thanks{}
%   $^1$School of Computing, University of Georgia \\ $^2$Department of Marine Sciences, University of Georgia \\
%   $^3$Department of Geography, University of Georgia\\
%   $^4$Department of Geography and the Environment, University of Texas at Austin\\
%   \texttt{\{fukun.liu, atgreer, gengchen.mai25, jinsun\}@uga.edu}
% }

\author{Fukun Liu}
\email{fukun.liu@uga.edu}
\orcid{0009-0003-7441-8776}
\affiliation{%
 \institution{University of Georgia}
 \city{Athens}
 \state{Georgia}
 \country{USA}
}

\author{Adam T. Greer}
\email{atgreer@uga.edu}
\orcid{0000-0001-6297-9371}
\affiliation{%
 \institution{University of Georgia}
 \city{Athens}
 \state{Georgia}
 \country{USA}
}

\author{Gengchen Mai}
\email{gengchen.mai@austin.utexas.edu}
\orcid{0000-0002-7818-7309}
\affiliation{
 \institution{University of Texas at Austin}
 \city{Austin}
 \state{Texas}
 \country{USA}
}
\affiliation{%
 \institution{University of Georgia}
 \city{Athens}
 \state{Georgia}
 \country{USA} 
 }

\author{Jin Sun}
\email{jinsun@uga.edu}
\orcid{0009-0004-2926-4023}
\affiliation{%
 \institution{University of Georgia}
 \city{Athens}
 \state{Georgia}
 \country{USA}
}

\renewcommand{\shortauthors}{Fukun Liu, Adam T. Greer, Gengchen Mai and Jin Sun}
%%
%% The abstract is a short summary of the work to be presented in the
%% article.
\begin{abstract}
Plankton are small drifting organisms found throughout the world’s oceans and can be indicators of ocean health. One component of this plankton community is the zooplankton, which includes gelatinous animals and crustaceans (e.g. shrimp), as well as the early life stages (i.e., eggs and larvae) of many commercially important fishes. Being able to monitor zooplankton abundances accurately and understand how populations change in relation to ocean conditions is invaluable to marine science research, with important implications for future marine seafood productivity. 
While new imaging technologies generate massive amounts of video data of zooplankton, analyzing them using general-purpose computer vision tools %developed for general objects 
turns out to be highly challenging due to the high similarity in appearance between the zooplankton and its background (e.g., marine snow). In this work, we present the \textbf{ZooplanktonBench}, a benchmark dataset containing images and videos of zooplankton associated with rich geospatial metadata (e.g., geographic coordinates, depth, etc.) in various water ecosystems. ZooplanktonBench defines a collection of tasks to detect, classify, and track zooplankton in challenging settings, including highly cluttered environments, living vs non-living classification, objects with similar shapes, and relatively small objects.Our dataset presents unique challenges and opportunities for state-of-the-art computer vision systems to evolve and improve visual understanding in dynamic environments characterized by significant variation and the need for \textit{geo-awareness}. The code and settings described in this paper can be found on our website: \url{https://lfk118.github.io/ZooplanktonBench_Webpage}.
\end{abstract}

\begin{CCSXML}
<ccs2012>
   <concept>
       <concept_id>10010147.10010178.10010224.10010245.10010251</concept_id>
       <concept_desc>Computing methodologies~Object recognition</concept_desc>
       <concept_significance>500</concept_significance>
       </concept>
   <concept>
       <concept_id>10010147.10010178.10010224.10010245.10010250</concept_id>
       <concept_desc>Computing methodologies~Object detection</concept_desc>
       <concept_significance>500</concept_significance>
       </concept>
   <concept>
       <concept_id>10010147.10010178.10010224.10010245.10010253</concept_id>
       <concept_desc>Computing methodologies~Tracking</concept_desc>
       <concept_significance>300</concept_significance>
       </concept>
   <concept>
       <concept_id>10010405.10010432.10010437.10010438</concept_id>
       <concept_desc>Applied computing~Environmental sciences</concept_desc>
       <concept_significance>500</concept_significance>
       </concept>
 </ccs2012>
\end{CCSXML}

\ccsdesc[500]{Computing methodologies~Object recognition}
\ccsdesc[500]{Computing methodologies~Object detection}
\ccsdesc[300]{Computing methodologies~Tracking}
\ccsdesc[500]{Applied computing~Environmental sciences}

\keywords{Computer vision, Object detection, Fine-grained classification, Video analysis, Marine science}
\maketitle
%%%%%%%%% BODY TEXT
\vspace{-0.2cm}
\section{Introduction}\label{sec:introduction}
% \vspace{-0.35cm}

Plankton are small drifting organisms found throughout the world's oceans . One component of this plankton community is the zooplankton, which includes gelatinous animals and crustaceans (e.g. shrimp), as well as the young stages of many commercially important fishes. Because zooplankton are a critical component of marine food webs, there is a strong scientific motivation to monitor their abundances accurately and understand how their abundances change in relation to ocean conditions which is important for future seafood productivity prediction.

Imaging technologies are used to measure zooplankton abundances in situ with high spatial resolution, yet these technologies have also introduced many challenges \cite{benfield2007rapid,lombard2019globally}. In productive shallow water ecosystems (<200 m depth, where most commercial and recreational fishing occurs), zooplankton often live within dense patches of detrital material, also known as ``marine snow'' due to its fluffy or stringy appearance and gradual sinking speed. This marine snow includes dead and decaying phytoplankton (plankton that performs photosynthesis similar to terrestrial green plants), zooplankton fecal material, organic mucus, or other excretions from organisms \cite{turner2015zooplankton}. In situ imaging technologies have demonstrated that marine snow is highly patchy and can also have a variety of appearances making it difficult to differentiate zooplankton from marine snow. In addition, zooplankton have a variety of morphologies.
% that can be highly imbalanced.
% in terms of abundance. 
The young fishes and other larval animals are especially uncommon yet critical to identify accurately 
within the sea of marine snow common to biologically productive and economically important habitats.

These characteristics set up a challenge for machine vision in the ocean. \textit{How can we accurately separate living organisms from the much more common and variable marine snow?} 
% Once that is accomplished, 
\textit{Can we accurately identify rarer organisms like larval fishes and more common groups simultaneously?} Sometimes organisms within the images are interacting, and detecting those interactions automatically may be important for understanding zooplankton ecology \cite{greer2021spatial}. Accomplishing these goals would greatly enhance the ability to measure and understand ocean plankton populations. Solving these problems would have other applications for large imaging datasets where targets exist in environments with variable noise. Creative approaches are needed to combine available datasets (human annotated) and videos of live animals to accomplish these tasks.

In this paper, we propose \textbf{ZooplanktonBench} dataset that features a rich image and video dataset of zooplankton associated with geospatial metadata in various water ecosystems. Challenging benchmark tasks and tracks are designed to recognize zooplankton in complex environments. Our dataset presents unique challenges and opportunities for state-of-the-art computer vision systems. We expect our data can be beneficial to both the marine science and machine learning communities.

% \vspace{-0.35cm}
% \section{Related work}
% % \vspace{-0.35cm}

% Efforts to automatically identify plankton using machine vision has been underway for over 30 years \cite{hu2005automatic,sosik2007automated}, with more recent efforts gravitating towards the use of Convolutional Neural Networks (CNNs) \cite{luo2018automated,schmid2023edge}. Ellen et al. 2019 \cite{ellen2019improving} found that CNNs performed better when the plankton images were coupled with ``context metadata,'' which includes information from oceanographic sensors (i.e., the depth, temperature, salinity, and other ocean variables associated with each image). These results seem reasonable given that marine animals are patchy \cite{robinson2021big,greer2016examining} , meaning that one type is detected by an imaging system, and similar types are nearby that are also similar in size. Ideally, classification should rely solely on the image itself, just as an ocean researcher looks at specimens to make identifications (without bias as to where the organism came from). Nevertheless, the environmental origin of the training data could be relevant for creating an accurate classification algorithm. 
% % For example, can training data from one area be used to classify organisms at another depth?

\section{Related Work} \label{sec:related}

Efforts to automatically identify plankton using machine vision have been underway for over 30 years \cite{hu2005automatic,sosik2007automated}, with more recent efforts gravitating towards the use of Convolutional Neural Networks (CNNs) \cite{luo2018automated,schmid2023edge}. Ellen et al. \cite{ellen2019improving} found that CNNs performed better when the plankton images were coupled with ``context metadata,'' which includes information from oceanographic sensors (i.e., the depth, temperature, salinity, and other ocean variables associated with each image). These results seem reasonable given that marine animals are patchy \cite{robinson2021big,greer2016examining}, meaning that one type is detected by an imaging system, and similar types are nearby that are also similar in size. Ideally, classification should rely solely on the image itself, just as an ocean researcher looks at specimens to make identifications (without bias as to where the organism came from). Nevertheless, the environmental origin of the training data could be important. 
% relevant for creating an accurate classification algorithm. 

% \vspace{-0.35cm}
\section{\dataname{} Data}\label{sec:data}
% \vspace{-0.35cm}

Our \dataname{} dataset is collected from the northern Gulf of Mexico (nGOM), which is a highly biologically productive system that fuels fisheries and coastal economies in the United States. One reason 
for that high biological productivity 
is the high amount of river discharge that enters into the nGOM carrying high amounts of nutrients. These nutrients fuel the growth of photosynthesizing phytoplankton, which serve as the food for some types of zooplankton, and that energy eventually makes it up to productive fisheries \cite{grimes2001fishery}. The earliest stages of fishes are known as larvae, and they are 1 mm to several cm in size and are considered members of the zooplankton community. This study took place underneath a river plume, so the images from the shallowest depths (10 m) are just underneath the plume and feature large amounts of dead and decaying phytoplankton, known as marine snow. At the deeper depths, 25m and 35m, the images have less marine snow because much of the snow has been consumed or broken down.% by those depths. 

\begin{figure}[ht!]
	\centering \small %\tiny
	\begin{subfigure}[b]{0.165\textwidth} 
		\centering 
		\includegraphics[width=\textwidth]{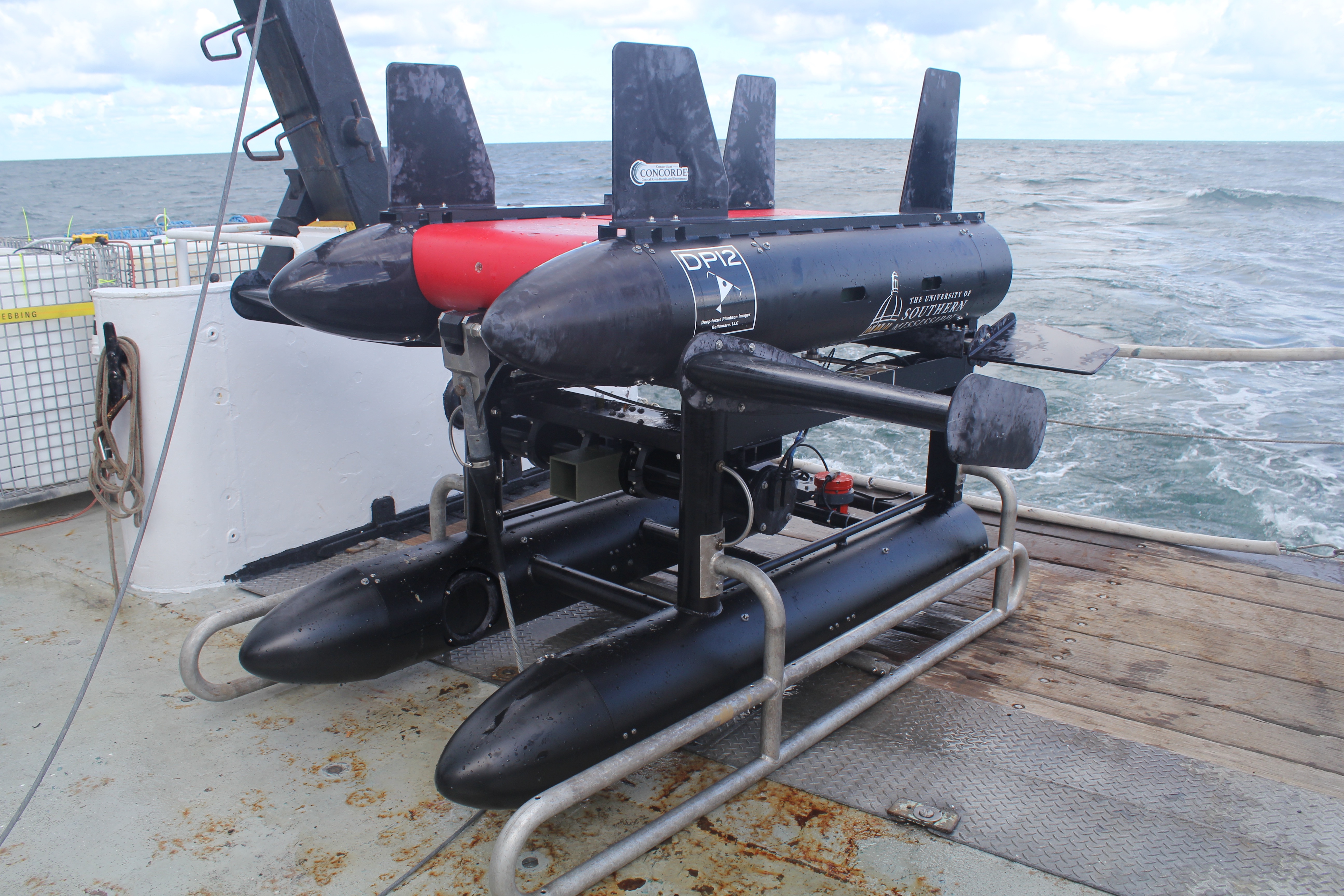}
		\caption[]%
		{{ISIIS
		}}    
		\label{fig:ISIIS}
	\end{subfigure}
	\hfill
	\begin{subfigure}[b]{0.095\textwidth} 
		\centering 
		\includegraphics[width=\textwidth]{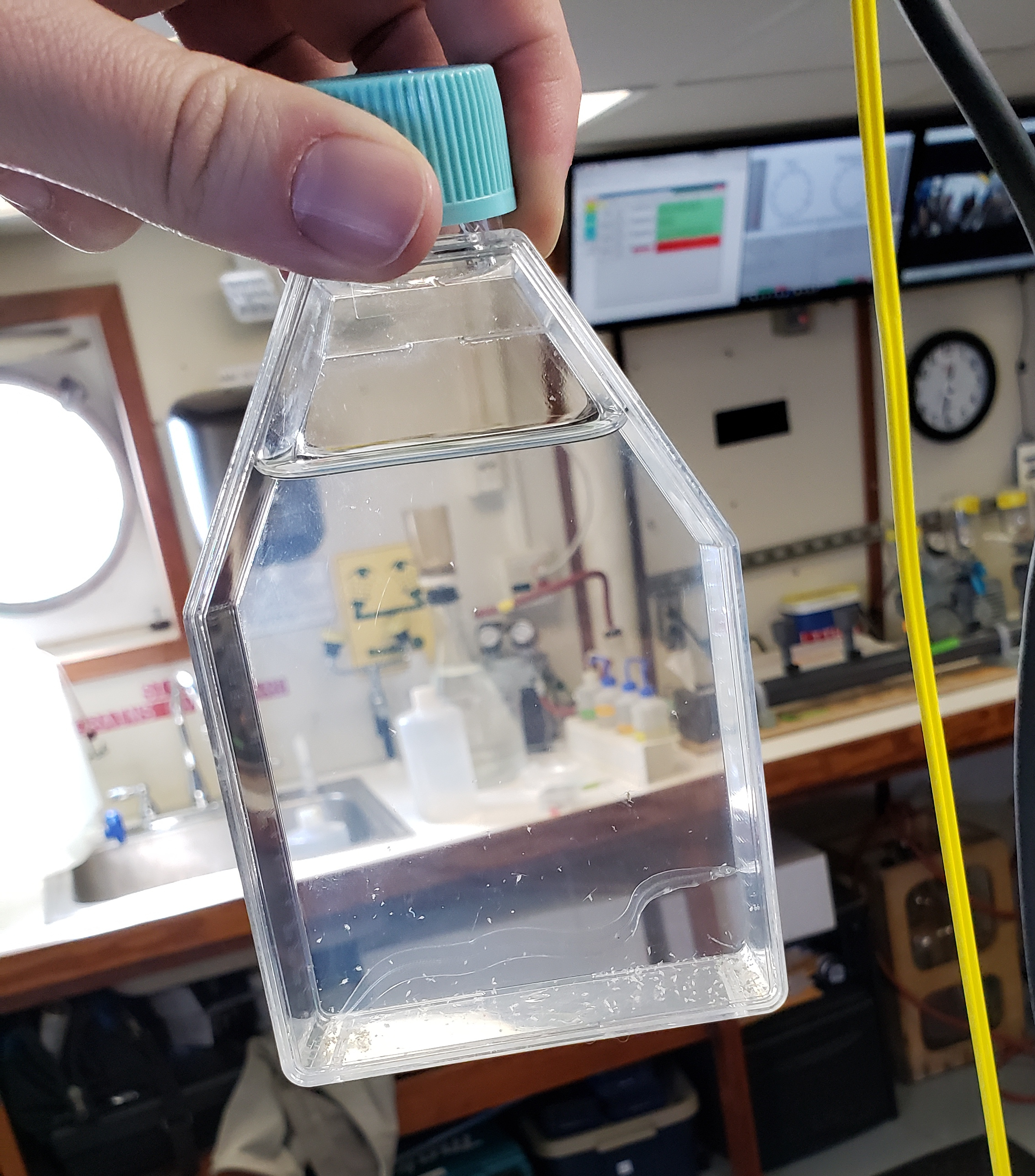}
		\caption[]%
		{{Eel larva
		}}    
		\label{fig:eel_larve_exp}
	\end{subfigure}
        \hfill
	\begin{subfigure}[b]{0.195\textwidth}  
		\centering 
		\includegraphics[width=\textwidth]{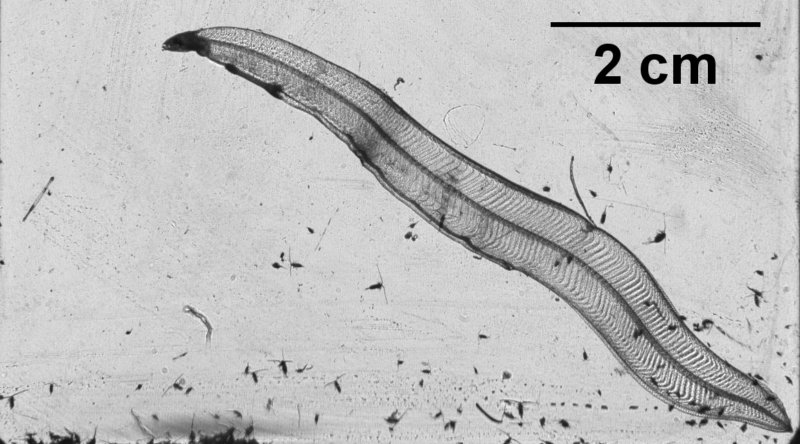}
		\caption[]%
		{{Eel larva image
		}}    
		\label{fig:eel_larve_img}
	\end{subfigure}
 \vspace{-0.4cm}
	\caption{
    (a) The In Situ Ichthyoplankton Imaging System (ISIIS, left) on the deck of a research vessel. Plankton or particles that pass in between the bottom two pods are captured in the images as "shadows." (b) An example of an eel larva (Leptocephalus larva) captured in a container. (c) The same eel larva as viewed in a shadowgraph imaging system.
	} 
	\label{fig:device}
	\vspace*{-0.3cm}
\end{figure}

\begin{figure}[h]
    \centering
    \includegraphics[width=1\linewidth]{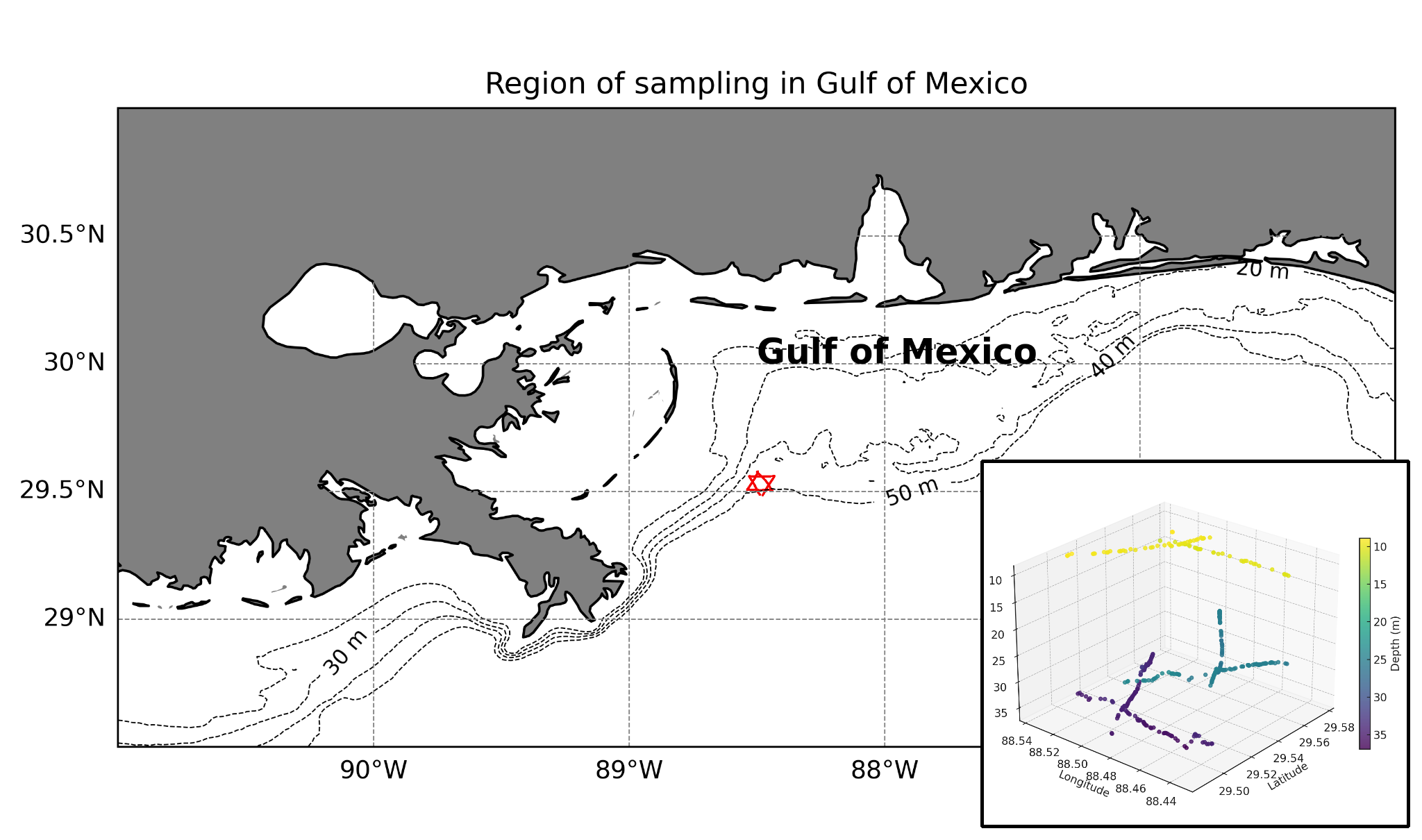}
    \vspace{-0.6cm}
    \caption{The region of sampling in the nGOM. Bathymetry contours correspond to 10-m intervals (20, 30, 40, and 50 m contours are labeled).
    The red line corresponds to the star-shaped sampling pattern between the 40 and 50-m isobaths, including crossings at three different depths (10, 25, and 35 m).
    The bottom right inset provides a 3D view of the sampling region.}
    \vspace{-0.4cm}
    \label{fig:geo-map}
    % \vspace{-0.4cm}
\end{figure}

Our data was collected with a towed plankton imaging vehicle,
the In Situ Ichthyoplankton Imaging System (see Figure \ref{fig:ISIIS}), %(ISIIS, Cowen and Guigand 2008)
on July 24, 2011 
% This vehicle, known as the In Situ Ichthyoplankton Imaging System \ref{fig:device} %(ISIIS, Cowen and Guigand 2008)
\cite{cowen2008situ}. It was towed behind a ship at 3 fixed depths in nGOM just east of the Louisiana Bird Foot Delta %(Greer et al. 2016)
\cite{greer2016examining} (see Figure \ref{fig:geo-map}). The ISIIS uses a small point light source, two plano-convex lenses, and a line-scan camera to collect images at 16Hz \cite{cowen2008situ}. The vehicle is towed at a constant speed of 2.5 meters per second 
% and uses motor-actuated wings to move the towed vehicle up and down in the water column, or the wings can be removed to 
and
easily maintain a constant depth (as was done in this study). The images are associated with oceanographic sensor data, such as geographic coordinates, depth, temperature, salinity, dissolved oxygen, and chlorophyll-a fluorescence (which serves as a proxy for phytoplankton abundance). These datasets provide a high-resolution map of oceanographic properties and how the organisms respond to these conditions.
The images are collected with the bottom two pods that house a camera and a light source. The marine creatures and marine snow passing between the pods block the light and are photographed as a ``shadow''.
Images were transferred via a fiber optic cable to a shipboard computer where they were stored on hard drives. The system has real-time information on the vehicle position, the associated oceanographic sensor data, and live-streaming data. Figure \ref{fig:eel_larve_exp} and \ref{fig:eel_larve_img} show an example of an eel larva (Leptocephalus larva) captured in a container as well as its image view in a shadowgraph imaging system. 
% along with live streaming of images.

\vspace{-0.2cm}
% \begin{figure}[h]
%     \centering
%     \includegraphics[height=1.3in]{Picture/device_1.jpg}    
%     \includegraphics[height=1.3in]{Picture/20210210_123557_rot_crop.png}    
%     \includegraphics[height=1.3in]{Picture/LeptoFeb.jpg}    
%     \caption{Top left: The In Situ Ichthyoplankton Imaging System (ISIIS, left) on the deck of a research vessel. Plankton or particles that pass in between the bottom two pods are captured in the images as "shadows." Top right: an example of an eel larva (Leptocephalus larva) captured in a container. Bottom: the same eel larva as viewed in a shadowgraph imaging system.}
%     \label{fig:device}
%     \vspace{-0.4cm}
% \end{figure}

\begin{figure}[ht!]
	\centering \small %\tiny
	\begin{subfigure}[b]{0.12\textwidth} 
		\centering 
		\includegraphics[height=0.5in]{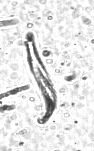}
		\caption[]%
		{{\small Chaetognath
		}}    
		\label{fig:chaeto_1}
	\end{subfigure}
	\hfill
	\begin{subfigure}[b]{0.12\textwidth} 
		\centering 
		\includegraphics[height=0.5in]{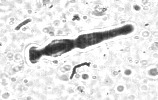}
		\caption[]%
		{{\small Chaetognath
		}}    
		\label{fig:chaeto_2}
	\end{subfigure}
        \hfill
	\begin{subfigure}[b]{0.12\textwidth} 
		\centering 
		\includegraphics[height=0.5in]{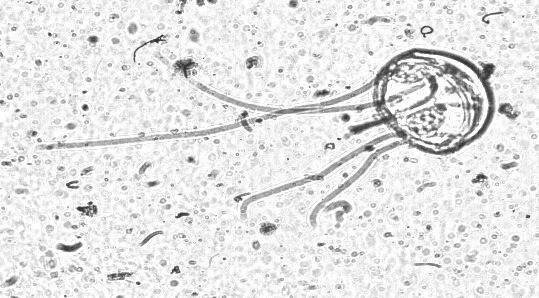}
		\caption[]%
		{{\small Hydromedusa
		}}    
		\label{fig:Hydro_1}
	\end{subfigure}
	\hfill
	\begin{subfigure}[b]{0.15\textwidth} 
		\centering 
		\includegraphics[height=0.5in]{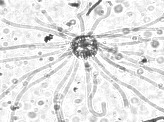}
		\caption[]%
		{{\small Hydromedusa
		}}    
		\label{fig:Hydro_2}
	\end{subfigure}
        \hfill
        \begin{subfigure}[b]{0.15\textwidth} 
		\centering 
		\includegraphics[height=0.5in]{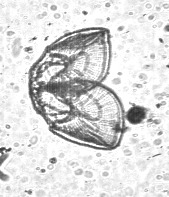}
		\caption[]%
		{{\small Lobate ctenophore
		}}    
		\label{fig:LC_1}
	\end{subfigure}
	\hfill
	\begin{subfigure}[b]{0.15\textwidth} 
		\centering 
		\includegraphics[height=0.5in]{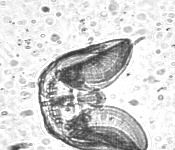}
		\caption[]%
		{{\small Lobate ctenophore
		}}    
		\label{fig:LC_2}
	\end{subfigure}
        \hfill
        \begin{subfigure}[b]{0.12\textwidth} 
		\centering 
		\includegraphics[height=0.5in]{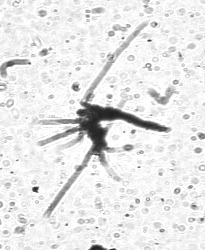}
		\caption[]%
		{{\small Shrimp
		}}    
		\label{fig:shrimp_1}
	\end{subfigure}
	\hfill
	\begin{subfigure}[b]{0.12\textwidth} 
		\centering 
		\includegraphics[height=0.5in]{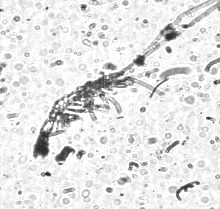}
		\caption[]%
		{{\small Shrimp
		}}    
		\label{fig:shrimp_2}
	\end{subfigure}
        \hfill
        \begin{subfigure}[b]{0.12\textwidth} 
		\centering 
		\includegraphics[height=0.5in]{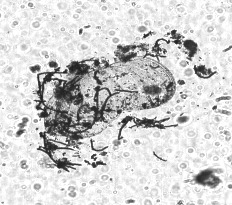}
		\caption[]%
		{{\small Unknown
		}}    
		\label{fig:Unk_1}
	\end{subfigure}
 \vspace{-0.4cm}
	\caption{
    Example photos of zooplankton in our \dataname{} dataset.
	} 
	\label{fig:zooplankton_exp}
	\vspace*{-0.5cm}
\end{figure}

Images from this dataset were processed by zooplankton experts. The images first underwent an automated flat-fielding procedure to remove artifacts and improve contrast. Then all identifiable organisms were extracted using customized keyboard shortcuts implemented in ImageJ %(Schneider et al. 2012)
\cite{schneider2012nih}. The 17 broad zooplankton identification categories used in this study and the statistics of the corresponding image annotations from different depths (e.g., 10m, 25m, and 35 m) are listed in Table \ref{tab:Dataset_summary}. 
% included chaetognaths, cumaceans, doliolids/salps (thaliaceans), miscellaneous ctenophores, ctenophores with tentacles, lobate ctenophores, fish larvae, hydromedusae, shrimp, polychaetes, pteropods, siphonophores, stomatopod larvae, and unknown plankton.
The unknown category includes rare taxa that did not fall into the specified categories and was not analyzed further. Crab zooea and copepods were only quantified on the deepest transects (35 m) because large individuals were not found in the shallower depths. All fish larvae were further classified to the family level, and all image classifications were checked and corrected. \textbf{A total of 374,378 images were analyzed, which took $\sim$800 person-hours to complete and generated in 285,733 cropped and identified organisms of interest}. Smaller and more abundant organisms, such as copepods and appendicularians, were present in almost every image but were not identified because it would have greatly increased the amount of time needed to classify each image. Thus, many living zooplankton were not labeled, and the identified organisms mostly belonged to larger sizes (i.e., between $\sim$5 mm and several centimeters in length). Figure \ref{fig:zooplankton_exp} shows some examples of zooplankton in our \dataname{} dataset. 
% \begin{figure*}[t]
%   \centering
%   \begin{tabular}{ccccc}
%   % {m{3cm}m{3cm}m{3cm}}
%   Chaetognath & Chaetognath & \multicolumn{2}{c}{Hydromedusa}& Hydromedusa\\
%   \includegraphics[height=0.7in]{Picture/chaeto_1.jpg} &
%   \includegraphics[height=0.7in]{Picture/chaeto_2.jpg} &
%   \multicolumn{2}{c}{\includegraphics[height=0.7in]{Picture/Hydro_1.jpg}} &
%   \includegraphics[height=0.7in]{Picture/Hydro_2.jpg} \\ 
%   Lobate ctenophore &Lobate ctenophore & Shrimp & Shrimp & Unknown\\
%   \includegraphics[height=0.7in]{Picture/LC_1.jpg} &
%   \includegraphics[height=0.7in]{Picture/LC_2.jpg} &
%   \includegraphics[height=0.7in]{Picture/shrimp_1.jpg} &
%   \includegraphics[height=0.7in]{Picture/shrimp_2.jpg} &
%   \includegraphics[height=0.7in]{Picture/Unk_1.jpg}
%   \end{tabular}
%   \caption{Example photos of zooplankton in our \dataname{} dataset.}
%   \label{plankton examples}
%   \vspace{-0.5cm}
% \end{figure*}

\begin{table*}[ht!]
  \caption{Detailed annotation information of \dataname.}
  \label{tab:Dataset_summary}
  \centering
  \resizebox{0.7\textwidth}{!}{ %\setlength{\tabcolsep}{1pt} 
  % \scriptsize
  \begin{tabular}{lcccc}
    \toprule
    Zooplankton Class& \makecell{Number of \\annotations}& \makecell{Annotation from \\ 10 meters}  & \makecell{Annotation from \\ 25 meters} & \makecell{Annotation from \\ 35 meters}\\
    \midrule
    Chaetognath or Arrow worm & 235,065 & 66,874 & 49,030 & 119,161\\
    Shrimp & 17,243 & 7,656 & 7,286 & 2,301 \\
    Fish larva & 2,432 & 1,011 & 772 & 649 \\
    Scyphomedusa & 856 & 571 & 269 & 16 \\
    Pleurobrachia & 823 & 606 & 138 & 79 \\
    Hydromedusa or Jellyfish & 7,821 & 2,033 & 983 & 4,805 \\
    Thaliacea & 452 & 81 & 279 & 91 \\
    Stomatopod larva & 1,024 & 474 & 502 & 48 \\
    Lobate ctenophore & 671 & 343 & 153 & 175 \\
    Siphonophore & 2994 & 774 & 948 & 1,272 \\
    Polychaete worm & 852 & 14 & 66 & 772 \\
    Ctenophore or Comb jelly & 22 & 10 & 10 & 2 \\
    Pteropod & 82 & 3 & 68 & 11 \\
    Cumacean & 2,418 & 0 & 0 & 2418 \\
    Crab zooea or Crab larva & 1,336 & 0 & 0 & 1,336 \\
    Copepod & 7,366 & 0 & 356 & 7,010 \\
    Unknown & 4,276 & 884 & 1,703 & 1,689 \\
    \midrule 
    Total & 285,733 & 81,334 & 62,564 & 141,835 \\

    \bottomrule
  \end{tabular}
  }
\end{table*}

In addition to the towed imaging dataset, videos were collected of organisms in some categories onboard a ship using a benchtop imaging system %(Greer et al. in review)
\cite{https://doi.org/10.1002/lom3.10657}. Live animals that we captured in a net and placed into a small container onboard a research vessel in preparation for filming. The videos feature zooplankton swimming around in a container at a frame rate of 25 Hz. The idea behind these videos was that they would be useful for quantifying swimming behaviors and building robust training libraries. The latter goal has yet to be evaluated but could be useful for augmenting training libraries and automatically detecting living animals in images (and separating them from marine snow).

This high-quality human-annotated dataset features a few other characteristics that make it valuable for computer vision. \textbf{The contrasting environments with different marine snow amounts are representative of a typical biologically productive environment.} So identifying the marine animals in these variable environments is an important task for understanding the connection of marine animals to these habitats, and the task is also representative of a general problem in computer vision in diverse and cluttered backgrounds. 
% As is typical with any marine animal image dataset, t
The identification categories are highly imbalanced (chaetognaths are very common, for example) and some categories, such as fish larvae, have many different morphologies because they belong to different species or stages of development. Together, the features of this dataset represent an opportunity to employ the latest techniques for object detection and recognition in variable backgrounds. On a practical level for ocean scientists, it is useful for living organisms to be separated from the marine snow, thus giving scientists a simpler dataset to annotate. If video is useful for improving image classification, that would be a major step forward for efficiency.
% in improving efficiency for zooplankton image processing.

% The In Situ Ichthyoplankton Imaging System %(ISIIS, Cowen and Guigand 2008)
% \cite{cowen2008situ} used to collect the benchmark dataset. 
% The images are collected with the bottom two pods that house a camera and a light source. The marine creatures and marine snow passing between the pods block the light and are photographed as a “shadow.”

% Images were transferred via a fiber optic cable to a shipboard computer where they were stored on hard drives for further analysis. The software provides real-time information on the vehicle position and the associated oceanographic sensor data, along with a live feed of the images.

% \vspace{-0.35cm}
\section{\dataname{} Tasks and Tracks}\label{sec:benchmark}
% \vspace{-0.35cm}

The imagery data collected in Section \ref{sec:data} presents rich geo-aware information and great challenges to computer vision systems. 
We design several tasks for \dataname, which focus on the unique characteristics of the data.
For each of the learning tasks, we make multiple tracks to train and evaluate computer vision models under different scenarios and setups.
In Section \ref{sec:baselines}, we show the performance of state-of-the-art CV systems on our data and point out their limitations and challenges for computer vision and machine learning.
Notably, the \dataname{} dataset 
% is not limited by the tasks we designed in this paper. It 
contains rich information that can be used for other computer vision tasks such as self-supervised video learning, or object tracking in complex scenes. We expect our published dataset can be found useful by those researchers as well.

\subsection{Tasks}\label{sec:tasks}
% Our primary objective is to detect and accurately categorize as much live plankton as possible from the data collected. However, the task presents significant challenges. We have obtained images from three different ocean depths: 10 meters, 25 meters, and 35 meters. Despite the plankton's appearance remaining consistent across these depths, the environmental complexity and variability in conditions at each depth complicate the detection and categorization process. Furthermore, plankton are inherently small, and individuals within the same species can exhibit considerable morphological differences, adding another layer of difficulty to consistent detection and categorization.
Our primary objective is to detect and classify zooplankton given imagery data collected in Section \ref{sec:data}.
We design the following learning-based tasks that are not only invaluable for marine science research but also challenging for the deep learning community. 

% \noindent\textbf{
\subsubsection{
Fine-grained detection and classification on zooplankton species}
As discussed in Section \ref{sec:introduction}, being able to correctly detect and classify zooplankton from images is invaluable to marine science. 
However, the appearance of the different zooplankton types is highly variable, making it a challenging task for learning systems.
In this task, we provide fine-grained zooplankton labels and their bounding boxes for each image and ask an object detection system to locate and classify them in new unseen images.
This fine-grained classification task is a fundamental problem in image classification and object detection in computer vision. 
We adopt a similar training setup and widely used evaluation metrics such as Mean Average Precision (50, 50-95) to measure the performance of various methods.

% \paragraph{Classification} Like a traditional classification task, we trained our labeled data with categories. Then, we apply the living creature results based on what we just mentioned to further filter the results. All the training data and validation data are from the real ocean data.

\begin{figure}[t]
    \centering
    \includegraphics[width=0.5\textwidth]{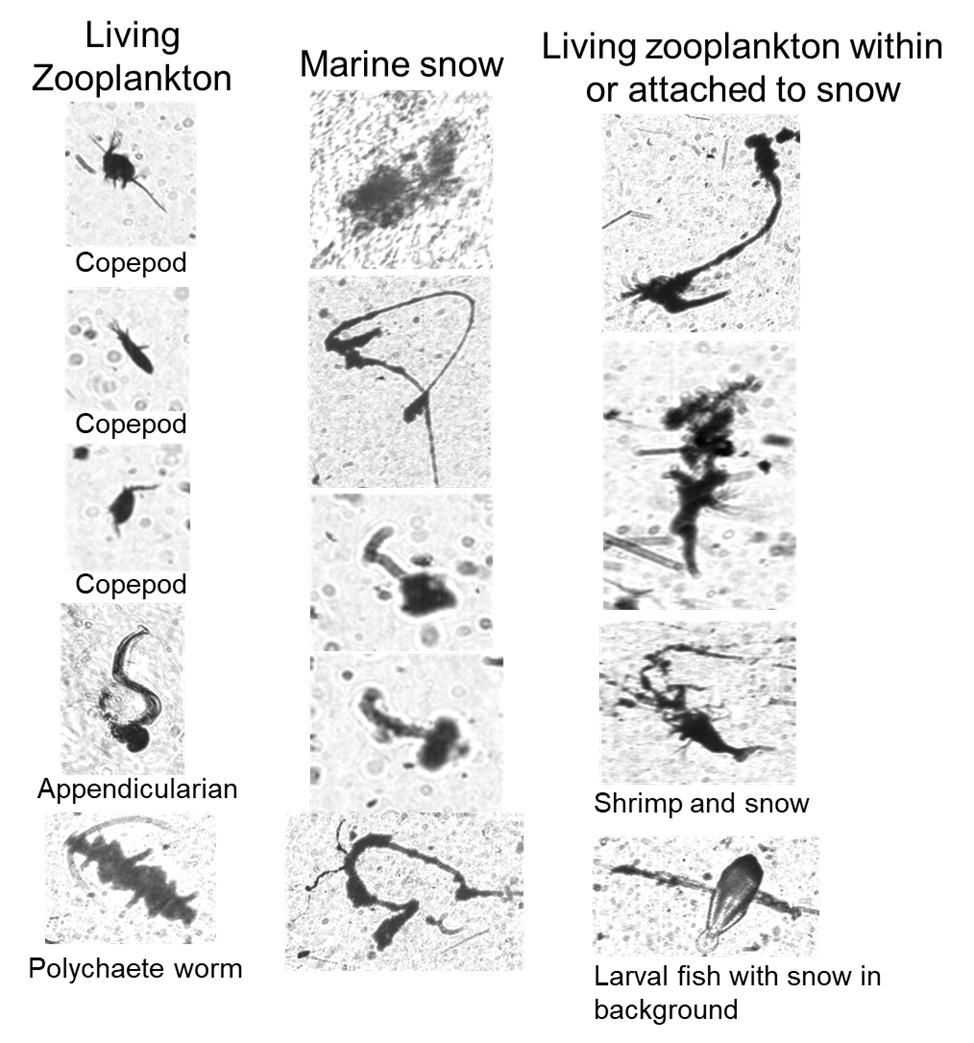}  \vspace{-0.3cm}
    \caption{Examples of zooplankton and marine snow. Marine snow is abundant and can take many different shapes. By chance, they can resemble living zooplankton in the images.}
    \label{fig:zooplankton-marine-snow}
    \vspace{-0.5cm}
\end{figure}

\subsubsection{Living zooplankton vs. marine snow detection}
One of the unique challenges in our data/task is the predominance existence of distractors (i.e., non-living marine snow particles) in the data. 
As shown in Figure \ref{fig:zooplankton-marine-snow}, there are a lot of objects that are visually similar to the zooplankton but they are actually belonging to ``marine snow.'' The high similarity in their appearances highlights the difficulty of differentiating living zooplankton from marine snow.
The marine snow consists of biological debris and organic material falling from upper waters to the deep ocean \cite{2020marine}. %\footnote{What is marine snow? National Ocean Service, NOAA. 11/05/2020.}. 
To make matters more challenging, in the video data, not only zooplankton can move in the water, the marine snow objects can also move in motion due to the disturbance of the water around them. 

In this task, we combine the fine-grained zooplankton species into one ``living zooplankton'' and essentially leave out all other distractors in the images as ``marine snow'' for this binary detection task. This is challenging because in many cases the differences between them are subtle, yet marine scientists can almost always easily distinguish living zooplankton from marine snow. Some objects, however, are ambiguous, so a marine scientist would classify them as "unknown" and cannot tell if the object is a marine snow particle or zooplankton. This only happens with relatively small objects ($\sim$10 pixels or $\sim$700 microns in length).

\subsection{Tracks}\label{sec:tracks}
\subsubsection{Diverse environments}
\begin{figure}[ht!]
	\centering \small %\tiny
	\begin{subfigure}[b]{0.23\textwidth} 
		\centering 
		\includegraphics[width=\textwidth]{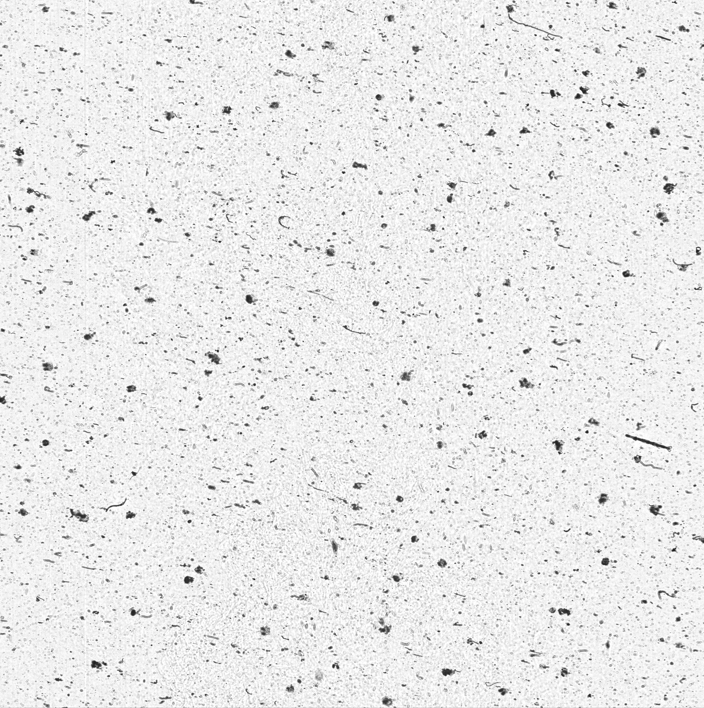}
		\caption[]%
		{{10 meters
		}}    
		\label{fig:10m_example}
	\end{subfigure}
	\hfill
	\begin{subfigure}[b]{0.23\textwidth} 
		\centering 
		\includegraphics[width=\textwidth]{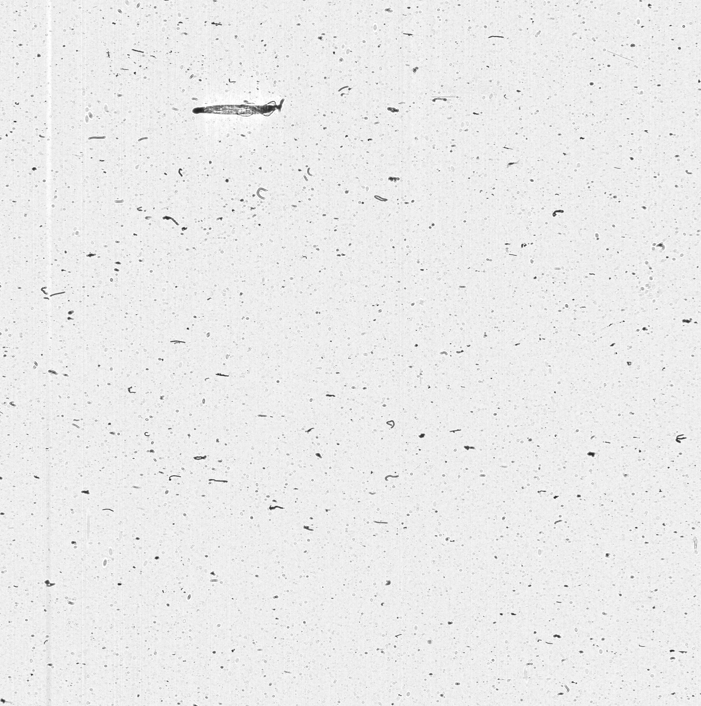}
		\caption[]%
		{{25 meters
		}}    
		\label{fig:25m_example}
	\end{subfigure}
        \hfill
	\begin{subfigure}[b]{0.23\textwidth}  
		\centering 
		\includegraphics[width=\textwidth]{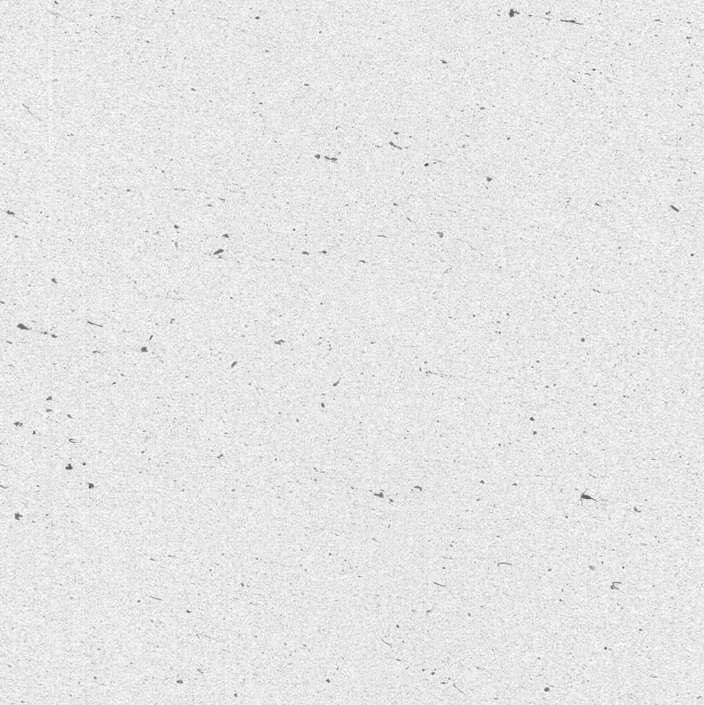}
		\caption[]%
		{{35 meters
		}}    
		\label{fig:35m_example}
	\end{subfigure}
 \vspace{-0.4cm}
	\caption{
    Image examples from 10 meters, 25 meters, and 35 meters. Each image is 13 cm by 13 cm and looks through 40 cm of ocean water (i.e., image depth of field).
	} 
	\label{fig:img_exp_depth}
	\vspace*{-0.4cm}
\end{figure}

As described in Section \ref{sec:data}, the samples we collected are from different depths in the ocean: 10 meters, 25 meters, and 35 meters. 
As a result, the characteristics of the data change significantly across the depths. 
Even though zooplankton appearance remains consistent across these depths, the imaging quality, environmental complexity, and variations pose great challenges to the detection and classification task.
Figure \ref{fig:img_exp_depth} shows example figures at the various depths. 

% \vspace{-0.2cm}
% \begin{figure}[t]
%     \centering
%     \includegraphics[width=0.22\textwidth]{Picture/10m_example.PNG}
%     \hfill
%     \includegraphics[width=0.22\textwidth]{Picture/25m_example.PNG}
%     \hfill
%     \includegraphics[width=0.22\textwidth]{Picture/35m_example.PNG}
%     \caption{From top to bottom are examples from 10 meters, 25 meters, and 35 meters. Each image is 13 cm by 13 cm and looks through 40 cm of ocean water (i.e., image depth of field).}
%     \label{fig:images from 3 different depths}
% \end{figure}

In this track setup, we evaluate the performance of a computer vision system on the tasks in Section \ref{sec:tasks} in diverse environments with differences in the background abundance of targets and marine snow. 
Specifically, we measure the classification and detection performance of a CV model that is trained on \{10 meters, 25 meters, 35 meters, all-mix\} datasets and tested on \{10 meters, 25 meters, 35 meters, and all-mix\} datasets.
This setup generates a comprehensive view of the accuracy and generalization ability of a model trained on the \dataname{} data, enabling an evaluation of its geo-aware performance with respect to ocean depth.

% \paragraph{Complexity of environment} The complexity of the environment varies significantly across different depths of ocean data, as depicted in Figure \ref{fig:images from 3 different depths}. It is evident that the deeper the image is captured, the fewer objects it contains. This observation presents a unique challenge: How does a detector perform in varying environmental conditions? Specifically, what will the performance be if a model, trained on the dataset from 10 meters, is tested on data from 25 meters? Addressing these questions is crucial for developing a model that can effectively adapt to and operate within diverse marine settings.

\subsubsection{Images + video}
Otherwise purely relaying on images (denoted as "Image Only"), it has been shown in prior research that training with a mix of fully supervised and unsupervised data can boost the performance of machine learning systems \cite{berthelot2019mixmatch}.
In particular, training with additional video data can help downstream single-image tasks \cite{parthasarathy2023self}. 
In our case, we augment the fully labeled image training set with the unlabeled video data. 

In this track, we evaluate the performance of a model on the tasks in Section \ref{sec:tasks} with the training of both the fully supervised image data and unlabeled video data on classification and detection. 
To simplify the preparation of the training setup, we provide an efficient object tracking tool, ByteTrack \cite{zhang2022bytetrack}, to work closely with any trained object detector on our data.
In Section \ref{sec:baselines}, we describe a basic pipeline to leverage video data in the training framework of a fine-tuned object detector. 

% \paragraph{Data enhancement by utilizing video} To enhance the robustness of our results and address potential imbalances in our dataset, we have integrated video data captured in a controlled tank environment, characterized by a higher density of plankton and reduced presence of marine snow. This modification aims to produce more accurate results tailored to specific research needs. By utilizing multiple object tracking techniques, specifically ByteTrack [insert citation here], in conjunction with our fine-tuned YOLO model, we generate a richer set of labeled images. We meticulously retain only those objects that are consistently recognized correctly throughout the tracking process, ensuring the robustness of the video-generated results. These images, derived from video sequences, provide diverse instances of live plankton. This controlled setting allows us to enrich our dataset substantially and enhances the model’s capacity to generalize across varied conditions. Consequently, this augmentation significantly improves the detection accuracy and reliability of the model, making it more adaptable to our specific requirements.

% \vspace{-0.35cm}
\section{Experiments}\label{sec:baselines}
% \vspace{-0.35cm}

Here we show the performance of representative state-of-the-art detection and classification computer vision algorithms trained and tested on our data with the tasks and tracks defined in Section \ref{sec:benchmark}.
For each method, we also discuss their limitations and challenges. All the experiments are conducted on a Linux Ubuntu workstation equipped with an NVIDIA RTX A6000 GPU with 48GB. The average training time for one fine-tuned model is 15 hours.

\subsection{Fine-tuned object detector}
YOLO (You Only Look Once) is a cutting-edge object detection model \cite{yolo} that has evolved over multiple iterations and gained widespread adoption in various applications due to its speed and accuracy \cite{lou2023dc,talaat2023improved,xiao2024fruit}.
In this work, we select the recently released YOLOv8 \cite{Jocher_Ultralytics_YOLO_2023} as the base object detector.
We fine-tune YOLOv8 for two tasks: fine-grained detection/classification and living zooplankton vs. marine snow detection.

For the image + video track, we integrated the fine-tuned YOLOv8 model with ByteTrack to produce \textit{object tracking} results. These results will include labeled objects, with some objects assigned an ``id'' and others without one. The ``label'' is provided by YOLOv8, while the ``id'' is assigned by ByteTrack. An object assigned an ``id'' indicates that it is trackable and consistently detected as the same class across multiple frames (5 frames in our experiment). This consistency suggests a higher likelihood of the object being a living zooplankton compared to objects without it.  Based on this assumption, we can enhance the detection accuracy by disregarding labeled objects that are untracked.
% For our project, we have selected YOLOv8 as the foundational model, which we intend to fine-tune using our dataset to meet specific detection needs. We have defined two primary tasks for the detector. The first task involves differentiating live plankton from "marine snow," which is the detrital material consisting of dead plankton. The second task is to classify different species of plankton. 

\begin{figure}[h]
    \centering
    \includegraphics[width=0.48\textwidth]{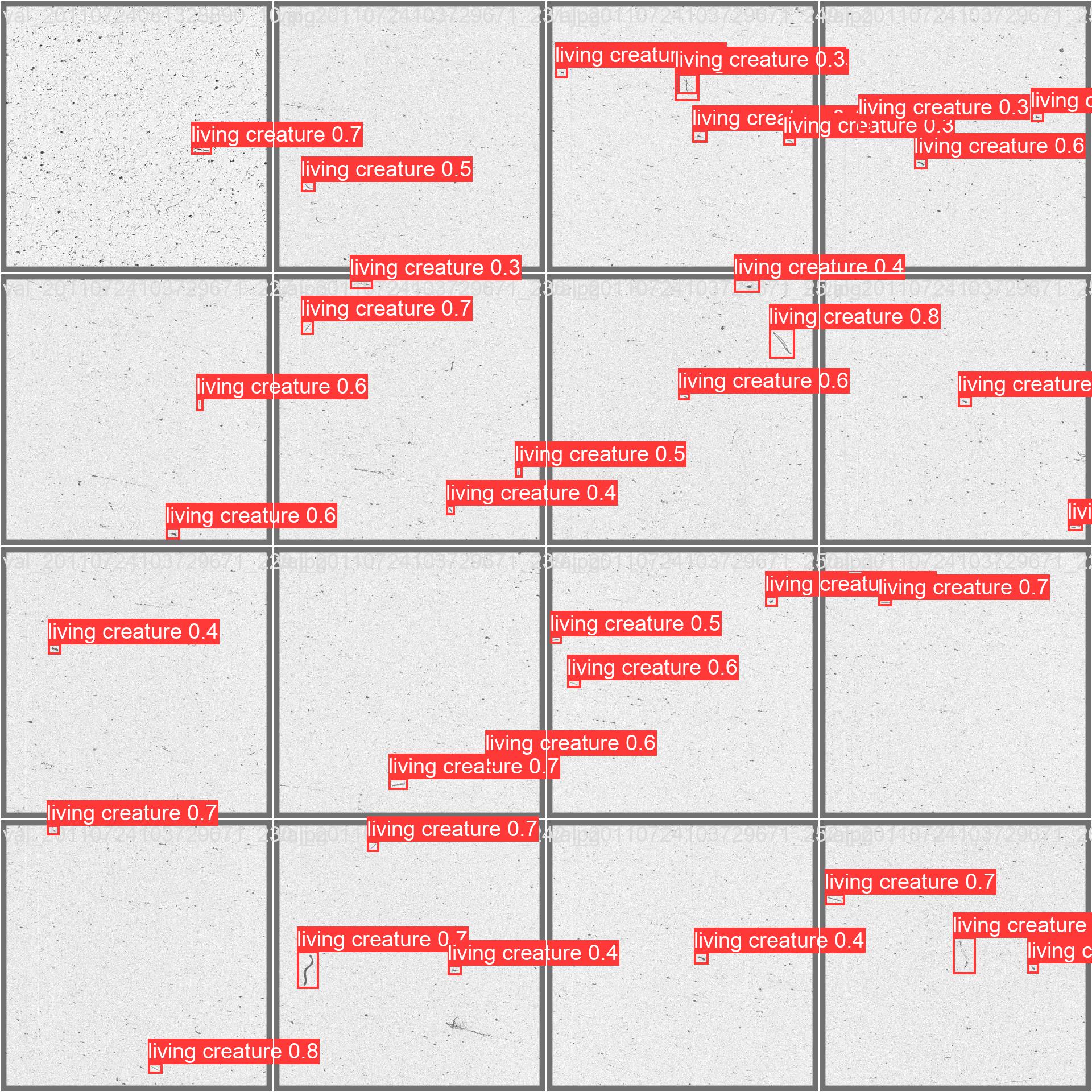}
    \includegraphics[width=0.48\textwidth]{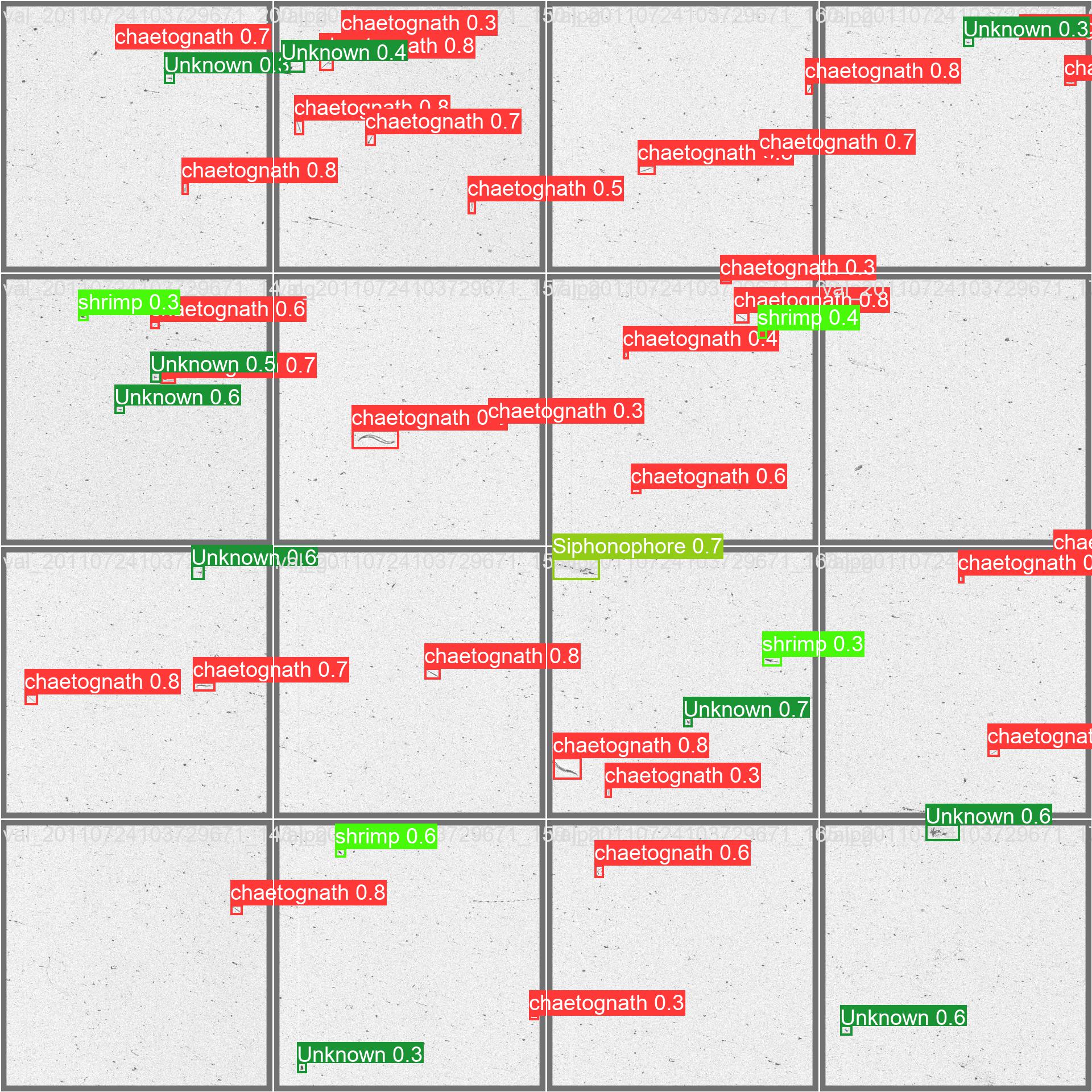}    
    \caption{\textbf{YOLOv8} detection and classification results. Top: living zooplankton detection. Bottom: fine-grained classification.}
    \label{fig:yolo-visual}
\end{figure}

\noindent\textbf{Results}.
Table \ref{tab:yolo summary} 
% shows the quantitative results of the YOLOv8's 
% fine-grained classification performance on \dataname.
shows YOLOv8's results on the fine-grained zooplankton detection and classification task under the diverse environments track. 
%The nomenclature "result\_N\_M" is used to denote the scenario where the model is trained using data from N meters and tested on data from M meters. 
% Additionally, "mix" refers to the combined dataset from all three depths—10 meters, 25 meters, and 35 meters. 
Table \ref{tab:live plankton - ocean video} shows YOLOv8's results on the living zooplankton vs. marine snow detection task under the regular image-only and the images + video track.
% Table \ref{tab:live plankton - video} shows YOLOv8's results on the same task under the images + video track.
Figure \ref{fig:yolo-visual} shows examples of detection results of living zooplankton and fine-grained classification results of zooplankton using the trained model.

% \begin{table}
%   \caption{Experimental results of YOLOv8 on the living zooplankton vs. marine snow detection task.}
%   \label{tab:live plankton - ocean}
%   \centering
%   \begin{tabular}{lll}
%     \toprule
%     Model      & mAP50  & mAP50-95 \\
%     \midrule
%     Fine-tuned YOLOv8         & 0.871  & 0.522  \\    
%     \bottomrule
%   \end{tabular}
% \end{table}

% \begin{table}
%   \caption{Experimental results of YOLOv8 on the living zooplankton vs. marine snow detection task under the images + video track.}
%   \label{tab:live plankton - video}
%   \centering
%   \begin{tabular}{lll}
%     \toprule
%     Model      & mAP50  & mAP50-95 \\
%     \midrule    
%         Images + video YOLOv8      & 0.84   & 0.475  \\
%     \bottomrule
%   \end{tabular}
% \vspace{-0.35cm}
% \end{table}

% \begin{figure}[h]
%     \centering
%     \includegraphics[width=0.6\textwidth]{Picture/YOLO_Living_example.jpg}    
%     \caption{Live zooplankton detection}
%     \label{fig:images for living}
% \end{figure}

% \begin{figure}[h]
%     \centering
%     \includegraphics[width=0.6\textwidth]{Picture/YOLO_Classification_example.jpg}    
%     \caption{Zooplankton classification}
%     \label{fig:images for classification}
% \end{figure}

\begin{table}
  \caption{Experimental results of \textbf{YOLOv8} on the fine-grained zooplankton species classification task under the diverse environment track. ``mix'' refers to the combined dataset from all three depths—10 meters, 25 meters, and 35 meters.}
  \label{tab:yolo summary}
  \centering
  \begin{tabular}{ll|ll}
    \toprule
    Training dataset  &Testing dataset       & mAP50  & mAP50-95 \\
    \midrule
    10 meters &10 meters    & 0.303    & 0.162\\
    10 meters &25 meters    & 0.328    & 0.164\\
    10 meters &35 meters   & 0.128    & 0.0547\\
    10 meters &mix data   & 0.193    & 0.0991\\ \hline
    25 meters &10 meters    & 0.182    & 0.0941\\
    25 meters &25 meters & 0.594    & 0.354\\
    25 meters &35 meters    & 0.179    & 0.0813\\
    25 meters &mix data  & 0.229    & 0.124\\ \hline
    35 meters &10 meters    & 0.0799   & 0.0389\\
    35 meters &25 meters    & 0.12     & 0.0738\\
    35 meters &35 meters   & 0.411    & 0.284\\
    35 meters &mix data   & 0.138    & 0.0839\\ \hline
    mix data &10 meters   & 0.328    & 0.173\\
    mix data &25 meters   & 0.619    & 0.358\\
    mix data &35 meters   & 0.486    & 0.252\\
    mix data &mix data  & 0.434    & 0.228\\
    \bottomrule
  \end{tabular}
  % \vspace{-0.2cm}
\end{table}

\begin{table}
  \caption{Experimental results of \textbf{YOLOv8} on the living zooplankton vs. marine snow detection task.}
  \label{tab:live plankton - ocean video}
  \centering
  \begin{tabular}{llll}
    \toprule
    Track & Model      & mAP50  & mAP50-95 \\
    \midrule
    Image Only & YOLOv8         & 0.871  & 0.522  \\    
%     \bottomrule
%   \end{tabular}
% \end{table}

% \begin{table}
%   \caption{Experimental results of YOLOv8 on the living zooplankton vs. marine snow detection task under the images + video track.}
%   \label{tab:live plankton - video}
%   \centering
%   \begin{tabular}{lll}
%     \toprule
%     Model      & mAP50  & mAP50-95 \\
    % \midrule    
        Images + video &  YOLOv8 + ByteTrack     & 0.84   & 0.475  \\
    \bottomrule
  \end{tabular}
% \vspace{-0.35cm}
\end{table}

\noindent\textbf{Discussion}.
Table \ref{tab:yolo summary} shows that, except for the model trained at 35 meters, all other models achieve the best results on the 25-meter dataset. As previously discussed in Section \ref{sec:tracks}, environmental complexity varies with depth. Images at the 25-meter depth are characterized by an environment that is relatively sparse (compared to 10 meters) and may have zooplankton composition most similar to the other 2 depths. In addition, 25 meters appears to be the most balanced environment for the YOLOv8 detector. Further examination of performance on the ``mix'' dataset reveals that the model trained on the 25 meters dataset outperforms those trained at 10 and 35 meters in mAP, thereby reaffirming the importance of the 25 meters data. 
% However, this does not imply that data from other depths are irrelevant. 
Together, the model trained on the mix depth data significantly outperforms the model trained solely at 25 meters across all test cases, suggesting that integrating data from various depths enhances the model’s accuracy and generalizability. A more effective model could be developed by carefully mixing data from different depths.

\noindent\textbf{Challenges}.
Fine-tuning an object detector on custom data is one of the most widely used computer vision applications.
While the YOLOv8 detectors show reasonable performance in fine-grained classification and detection, the absolute numbers in mAP50-95 are significantly worse than that of the general object detection results (such as those in the COCO benchmark \cite{coco}).
It is known that common general object detectors could suffer from objects with small scale and in cluttered backgrounds \cite{hoiem2012diagnosing,liu2020deep}. 
Therefore we expect the challenges exist in our tasks and tracks can help the community to design better object detection algorithms that can benefit the general object tasks.
% recognition and detection tasks.

\subsection{Open-set object detector}
Recently, open-set object detectors such as Grounding DINO \cite{liu2023grounding} have gained popularity largely due to their flexibility. 
Once trained, these detectors can output a detection result with an arbitrary object prompt that is potentially unseen during training. 
This contrasts with traditional classifiers that operate under the closed-set assumption—namely, that the test sample's ground truth label will belong to one of the categories pre-defined during the training phase.
We adopt the Grounding DINO model  % \textit{without training/fine-tuning} 
and evaluate its open-set detection performance on \dataname{} in a zero-shot manner.

% \paragraph{Open-set detection} An open-set detector is a specialized type of machine learning model or algorithm that is engineered to identify and appropriately manage instances of data that do not belong to any of the classes on which the model was trained. This contrasts with traditional classifiers that operate under the closed-set assumption—namely, that all test data will belong to one of the categories observed during the training phase.

% It is noteworthy that an open-set detector can operate without any training and still meet specific requirements; however, it also exhibits certain limitations:

\noindent\textbf{Results}.
We found that the confidence threshold for the open-set detectors is a critical factor in the final performance. This threshold must be adjusted based on the dataset, implying that a single threshold may not be universally effective across different scenarios. In our context, a single threshold setting proved inadequate for varying depths and categories within the marine environment.

Table \ref{tab:DINO_combined} shows the quantitative results of the Grounding DINO's zero-shot performance on \dataname{}'s zooplankton fine-grained classification task with different confidence thresholds. Table \ref{tab:DINO_live_combined} shows the zero-shot performance of the Grounding DINO on \dataname{}'s zooplankton detection task under different confidence thresholds.

% \begin{enumerate}
%     \item \textbf{Confidence Threshold Adjustment}: The first limitation involves setting an appropriate confidence threshold for detections. This threshold must be adjusted based on the dataset, implying that a single threshold may not be universally effective across different scenarios. In our context, a single threshold setting proved inadequate for varying depths and categories within the marine environment.
    
%     \item \textbf{Structural Shortcomings}: The second limitation stems from the detector’s inherent structure. To detect unseen inputs, the model incorporates a language decoder that assists in interpreting these inputs by categorizing them into known classes. Although this feature minimizes the need for extensive pre-training, it introduces challenges in accurate classification. Specifically, in our dataset, the detector struggles to distinguish between ``larva fish,'' ``shrimp,'' and ``chaetognath,'' possibly because the decoder erroneously groups these distinct inputs into similar or identical classes. As a result, the classification performance is considerably compromised. For example, as illustrated in Table \ref{tab:DINO_combined}, the model's performance metrics are disappointing: the mAP50 is only 0.2 and the mAP50-95 is 0.0429 on 10 meters' data with a confidence threshold of 0.1. In contrast, the YOLOv8 model, trained specifically for our dataset, significantly outperforms the open-set detector, achieving a mAP50 of 0.328 and a mAP50-95 of 0.173 on the same test dataset.
% \end{enumerate}

\begin{table*}
  \caption{The zero-shot performance of \textbf{Grounding DINO} on the zooplankton fine-grained classification task under different confidence thresholds.}
  \label{tab:DINO_combined}
  \centering
  \begin{tabular}{lllllll}
    \toprule
    Depth & \multicolumn{2}{c}{Threshold 0.1} & \multicolumn{2}{c}{Threshold 0.075} & \multicolumn{2}{c}{Threshold 0.05} \\
    \cmidrule(r){2-3} \cmidrule(lr){4-5} \cmidrule(l){6-7}
                  & mAP50  & mAP50-95 & mAP50  & mAP50-95 & mAP50  & mAP50-95 \\
    \midrule
    10 meters     & 0.2    & 0.0429   & 0.1809 & 0.0382   & 0.1229 & 0.0257   \\
    25 meters     & 0.0978 & 0.0212   & 0.0852 & 0.0189   & 0.0439 & 0.01     \\
    35 meters     & 0.0159 & 0.0051   & 0.0125 & 0.0039   & 0.0082 & 0.0023   \\
    \bottomrule
  \end{tabular}
  % \vspace{-0.2cm}
\end{table*}

\begin{table*}
  \caption{The zero-shot performance of \textbf{Grounding DINO} on living zooplankton detection task under different confidence thresholds.}
  \label{tab:DINO_live_combined}
  \centering
  \begin{tabular}{lllllll}
    \toprule
    Depth & \multicolumn{2}{c}{Threshold 0.1} & \multicolumn{2}{c}{Threshold 0.075} & \multicolumn{2}{c}{Threshold 0.05} \\
    \cmidrule(r){2-3} \cmidrule(lr){4-5} \cmidrule(l){6-7}
                  & mAP50  & mAP50-95 & mAP50  & mAP50-95 & mAP50  & mAP50-95 \\
    \midrule
    10 meters     & 0.1352 & 0.0312   & 0.1106 & 0.0251   & 0.0585 & 0.0138   \\
    25 meters     & 0.0628 & 0.0123   & 0.0208 & 0.009    & 0.0267 & 0.0053   \\
    35 meters     & 0.0274 & 0.0057   & 0.046  & 0.0045   & 0.0125 & 0.0027   \\
    \bottomrule
  \end{tabular}
  % \vspace{-0.5cm}
\end{table*}

\noindent\textbf{Discussion}.
% In open-set object detection, training data is not required to achieve our objectives. However, this does not imply that implementation is straightforward or that optimal results are instantly achievable. Instead, substantial effort must be dedicated to fine-tuning the confidence threshold. 
% Analysis of 
Table \ref{tab:DINO_combined} shows that a higher threshold generally correlates with improved performance, indicating a positive relationship between the threshold and the performance. Moreover, an examination of Table \ref{tab:DINO_live_combined} reveals that performance at 25 meters does not consistently decline with a reduced threshold. Although the overall trend aligns with that observed in Table \ref{tab:DINO_combined}, caution is still necessary. 
% It is crucial to adjust the confidence threshold based on the specific conditions of each dataset, as depending solely on general rules derived from other datasets can lead to suboptimal outcomes.
This reveals how sensitive such open-set detectors can be with the proper setup of the thresholds. Sometimes this could be a challenge to end users.
Besides the confidence threshold, the performance of Grounding DINO on different depth datasets from \dataname{} varies significantly. For instance, the mAP50 scores at 10-meter, 25-meter, and 35-meter datasets, with a threshold of 0.1, are 0.2, 0.0978, and 0.0159 respectively. The mAP decreases by 51.1\% when moving from 10 meters to 25 meters and by 83.7\% from 25 meters to 35 meters. 
% In contrast to YOLOv8, which performs better at 25 meters, Grounding DINO is notably more effective at 10 meters than at the other two depths.
Unlike YOLOv8 which shows the best performance at 25 meters, Grounding DINO is more effective at 10 meters. This shows the different characteristics of the object detectors.

\noindent\textbf{Challenges}.
Besides the sensitivity of the threshold discussed above, one major challenge we found about the open-set detector is the prompt-based detection design. 
To detect unseen inputs, the model incorporates a language decoder that assists in interpreting these inputs by categorizing them into known classes. Although this feature minimizes the need for extensive pre-training, it introduces challenges in accurate classification. Specifically, in our dataset, the detector struggles to distinguish between ``larval fish,'' ``shrimp,'' and ``chaetognath,'' possibly because the decoder erroneously groups these distinct inputs into similar or identical classes. As a result, the classification performance is considerably compromised. For example, as illustrated in Table \ref{tab:DINO_combined}, the model's performance metrics are disappointing. The mAP50 is only 0.2 and the mAP50-95 is 0.0429 on 10-meter data with a confidence threshold of 0.1. In contrast, the YOLOv8 model, trained specifically for our dataset, significantly outperforms the open-set detector, achieving a mAP50 of 0.328 and a mAP50-95 of 0.173 on the same test dataset.
We hope our data and benchmark can be used as a testbed to evaluate the generalization ability of future open-set object detectors. 
% We are also interested in the fine-tuning or adapter-based extension of a general open-set detector to our data.

%%%%%%%%%%%%%%%%%%%%%%%%%%%%%%%%%%%%%%%%%%%%%%%%%%%%%%%%%%%%%%%%%
%%%%%%%%%%%%%%%%%%%%%%%%%% start %%%%%%%%%%%%%%%%%%%%%%%%%%%%%%%%    
\subsection{Large visual-language model}
Inspired by the success of foundational models such as GPT \cite{gpt2}, large visual-language models have been developed and demonstrated impressive results on many visual reasoning tasks \cite{liu2023llava}.
We evaluate the performance of GPT-4V from Microsoft Azure OpenAI \cite{openai2024chatgpt} on our \textit{geo-aware} dataset \dataname. Unlike traditional approaches with YOLO or Grounding DINO, where the original images are directly processed for detection, we employ a different methodology with GPT-4V:
% Instead of submitting the original images to GPT-4V, 
we first run the object detector on the input image and crop it according to the detected box.
% input the detection results from an initial object detectors. 
This strategy allows GPT-4V to perform our fine-grained zooplankton classification tasks, thereby enhancing the robustness of the results from the object detectors without significant additional computational costs.

We design two \textit{prompts} to the GPT4-V model, the first prompt is as follows:
\textit{``You are a Marine Biologist. I am sending you an image containing zooplankton. The possible classes are \textbf{chaetognath, larval fish, hydromedusa, lobate ctenophore, pleurobrachia, shrimp, siphonophore, stomatopod larva, unknown, thaliac, polychaete worm, scyphomedusa, other ctenophore}. Please give me the most probable name shown by the given image in the following format: class name. Please just give a short answer and do not provide explanations for your choice.''}

% The second prompt incorporates geo-metadata and structured as follows:
We also design a second prompt to explicitly encode \textit{geo-metadata} to evaluate the performance of GPT-4V being geo-aware \cite{mac2019presence,mai2020space2vec,yang2022dynamic,mai2023csp,mai2023sphere2vec,cole2023spatial,wu2024torchspatial}:
\textit{``You are a Marine Biologist. I am sending you an image containing zooplankton. The image was captured at latitude \underline{\textbf{29.5063167}} and longitude \underline{\textbf{88.5215167}}, situated \underline{\textbf{34.5}} meters beneath the seafloor. The environmental conditions at this location were as follows: an irradiance of \underline{\textbf{1.70E-05}}, a water pressure of \underline{\textbf{34.73}} dbar, a temperature of \underline{\textbf{22.21}} degrees Celsius, a salinity of \underline{\textbf{36.36}} psu, and a dissolved oxygen level of \underline{\textbf{1.8678}} mg per L. Given these conditions, the possible classes of zooplankton are: \textbf{chaetognath, larval fish, hydromedusa, lobate ctenophore, pleurobrachia, shrimp, siphonophore, stomatopod larva, unknown, thaliac, polychaete worm, scyphomedusa, other ctenophore}. Please give me the most probable name shown by the given image in the following format: class name. Please just give a short answer and do not provide explanations for your choice.``}

The underlined part will be replaced with the corresponding geo-metadata.

The detected-and-cropped images are passed directly to GPT-4V immediately following the prompt. GPT-4V then classifies each image crop by responding with one of the predefined classes or stating "I can't assist you" if it cannot recognize the image. Any response that does not match the established categories is classified as an error message. 
% This method ensures that all inputs are categorized, even those that the model fails to identify accurately, thereby maintaining systematic processing across the dataset.

% \begin{table}
%   \caption{\textbf{GPT-4V} fine-grained classification results on \dataname{}. `Correct' indicates the number of images correctly classified. `Classification' indicates the number of images predicted by GPT-4V, and `Instances' shows the total number of images in each class in ground truth data.}
%   \label{tab:GPT-4V classification}
%   \centering
%   \begin{tabular}{llll}
%     \toprule
%     Class         & Correct  &Classification & Instances  \\
%     \midrule    
%     Chaetognath           & 374    & 1066   & 508    \\
%     Larval fish           & 62     & 82   & 704   \\
%     Hydromedusa           & 23     & 32   & 158   \\
%     Lobate ctenophore     & 13     & 85   & 18  \\
%     Pleurobrachia         & 0      & 0   & 44   \\
%     Shrimp                & 7      & 44   & 306    \\
%     Siphonophore          & 5      & 42   & 98   \\
%     Stomatopod larva      & 13     & 33   & 118    \\
%     Unknown               & 0      & 41   & 18        \\
%     Thaliac               & 0      & 0   & 16        \\
%     Ctenophore            & 0      & 0   & 9        \\
%     \bottomrule
%   \end{tabular}
%   % \vspace{-0.4cm}
% \end{table}

\begin{table*}
  \caption{\textbf{GPT-4V} fine-grained classification results on \dataname{}. `Correct' indicates the number of images correctly classified. `Classification' indicates the number of images predicted by GPT-4V, the superscript on `Correct' and `Classification' indicates the data sent to GPT-4V, and `Instances' shows the total number of images in each class in ground truth data.}
  \label{tab:GPT-4V classification}
  \centering
  \begin{tabular}{llllll}
    \toprule
    Class & Correct$^{Image}$ &Classification$^{Image}$ &Correct$^{Image+Geo}$ &Classification$^{Image+Geo}$& Instances \\
    \midrule    
    Chaetognath           & \textbf{587}    & 1533 & 423    & 1194  & 666    \\
    Larval fish           & 78     & 114  & \textbf{86}     & 137  & 432   \\
    Hydromedusa           & \textbf{36}     & 109  & 9     & 45  & 162   \\
    Lobate ctenophore     & 11     & 118  & \textbf{14}     & 146 & 18  \\
    Pleurobrachia         & 0      & 0   & 0      & 0 & 45   \\
    Shrimp                & 3      & 29  & 3      & 22 & 362    \\
    Siphonophore          & \textbf{9}      & 37  & 2      & 10  & 108   \\
    Stomatopod larva      & \textbf{11}     & 13  & 6     & 10   & 126    \\
    Unknown               & 0      & 6  & 0      & 15 & 21        \\
    Thaliac               & 0      & 0  & 0      & 0 & 17        \\    
    Polychaete worm       & 1      & 108 & \textbf{3}      & 254  & 6       \\ 
    Scyphomedusa          &19      & 20  &\textbf{20}      & 20  & 216    \\
    Other ctenophore      & \textbf{3}      & 17  & 0      & 12  & 9        \\
    \bottomrule
  \end{tabular}
  % \vspace{-0.4cm}
\end{table*}

\noindent\textbf{Results}. We selected a dataset of 2,188 representative zooplankton images from \dataname{} for GPT-4V. Out of the total, without the input of geo-metadata, GPT-4V successfully classified 758 images but encountered 84 error messages during the classification process.  
With the input of geo-metadata, GPT-4V successfully classified 566 images but encounter 323 error messages during the classification process. The results of this classification are detailed in Table \ref{tab:GPT-4V classification}.

\noindent\textbf{Discussion}.
For the fine-grained classification task, GPT-4V seems to struggle with the concept of zooplankton.
% does not have a good understanding of what zooplankton is. 
Unlike its impressive results for classifying common objects such as cats and dogs, classifying zooplankton requires domain knowledge in marine science. Our \dataname{} could be a great supplement to such a need. 
With the zooplankton images alone, GPT-4V does performance well. In Table \ref{tab:GPT-4V classification}, ``Chaetognath'' is identified in 70.1\% of the total images, which is nearly triple the actual number of real chaetognath images present. This discrepancy suggests that GPT-4V struggles to accurately distinguish chaetognath from other types of zooplankton. This misclassification might be attributed to the widespread distribution of chaetognaths relative to other classes, increasing their likelihood of being represented in the training dataset utilized by GPT-4V. Additionally, an examination of the `Correct' and `Instances' columns reveals a high ratio of correctly identified chaetognath images to their total instances. This suggests that while GPT-4V frequently misclassifies other zooplankton as chaetognath, it successfully recognizes a substantial number of actual chaetognath instances.

% When we send the geo-meta information to GPT-4V (The second prompt), trying to help it do the classification task. The significantly increased error messages indicate that the complexity of input information will impact the behavior of GPT-4V. The geo-meta information improves accuracy for ''Lobate ctenophore'' (+16.67\%) and ''Larval fish'' (+1.85\%), but ''Chaetognath'' (-24.62\%) suffers a major drop despite fewer false positives. While ''Siphonophore'' (-20 false positives) and ''Chaetognath'' (-175 false positives) show better precision, ''Lobate ctenophore'' (+25 false positives) and ''Larval fish'' (+15 false positives) have increased false detections. ''Polychaete worm'' is heavily over-detected (form 108 to 254 false positives) but still mostly incorrect, and ''Hydromedusa'' saw a sharp drop in correct detections (from 36 to 9 false positives). ''Pleurobrachia'' and ''Thaliac'' remain undetected, while the ''Unknown'' category has more detections but no improvement in correctness. Overall, the introduce of geo-metadata enhances the result in some cases but struggles with accuracy, particularly for ''Chaetognath'' and ''Hydromedusa''.

As mentioned earlier, incorporating geo-metadata into GPT-4V increases error messages from 84 to 323, suggesting that additional information may not necessarily improve the model's performance.
As shown in Table \ref{tab:GPT-4V classification}, compared to GPT-4V using images alone, geo-metadata improves accuracy for ``Lobate ctenophore'' by 16.67\% and ``Larval fish'' by 1.85\% but reduces accuracy for ``Chaetognath'' by 24.62\%, despite fewer false positives.
% ``Siphonophore'' and ``Chaetognath'' exhibit improved precision with 20 and 175 fewer false positives, respectively. 
However, false detections increase for ``Lobate ctenophore'' by 25 and ``Larval fish'' by 15. ``Polychaete worm'' is heavily over-detected, with false positives rising from 108 to 254. ``Pleurobrachia'' and ``Thaliac'' remain undetected, and the ``Unknown'' category shows more detections without improved accuracy.
Overall, geo-metadata enhances classification in some cases but reduces accuracy for ``Chaetognath'' and ``Hydromedusa''.

\noindent\textbf{Challenges}.
While it has often been discussed that large foundational models are ``running out of data'' to be trained on \cite{out-of-data}, our evaluation shows that general-trained large models can struggle on data that require domain knowledge and fine-grained classification.
We hope \dataname{} can be used to evaluate the generalization ability of large visual and language models, maybe with a knowledge base type of approach: for example, by resorting to the literature in marine science, the model can learn the appearance features of different zooplankton types.

% \vspace{-0.35cm}
\section{Conclusion}
% \vspace{-0.35cm}

We present \dataname{}, % dataset and benchmark, 
a carefully designed and curated geo-aware dataset that features challenging and unique computer vision tasks such as detection and fine-grained classification of marine science imagery data. 
Our data features zooplankton that are highly diverse in appearance, and their habitat can be replete with with somewhat similar-looking marine snow particles.
Detection and classification systems developed in \dataname{} will be useful in the marine science community and may have applications for advancing computer vision and deep learning systems.

Several tests and developments were not explored in this paper that may have value in terms of scientific advancements and societal relevance. One limitation of \dataname{} is that we do not provide data annotation for our benchtop video dataset. Future work of \dataname{} may involve using annotated still images to develop an AI-assisted video annotation tool to help with this process. Improving algorithms to address the computer vision challenges illustrated in this benchmark dataset will help with the automated monitoring of marine populations and developing action plans to protect these ecologically and economically valuable areas of the planet.

\section*{Acknowledgements}
This work was supported by funds from the University of Georgia Presidential Interdisciplinary Seed Grant. Gengchen Mai acknowledges support from the Microsoft Research Accelerate Foundation Models Academic Research (AFMR) Initiative. Adam Greer is supported by funding from the U.S. National Science Foundation (OCE-2244690 and OCE-2023133).

% \bibliographystyle{plain}
% \bibliography{reference.bib}

\bibliographystyle{ACM-Reference-Format}
\balance
\bibliography{reference}

\newpage

% \newpage
% \thispagestyle{empty}
% \setcounter{table}{0}

\end{document}